\documentclass[]{fairmeta}
\usepackage{lineno}
\usepackage{calc}
\usepackage{amsmath}
\usepackage{caption}
\usepackage{multirow}
\usepackage{amsfonts}
\usepackage{xspace}
\usepackage[font=footnotesize]{subcaption}
\usepackage{booktabs}
\usepackage{pdfpages}
\usepackage{xcolor}
\usepackage[most]{tcolorbox}    
\usepackage{colortbl}
\usepackage{tabularx}
\usepackage{bbm}
\usepackage{graphicx}
\usepackage{algorithm}
\usepackage{algorithmic}
\usepackage{enumitem}
\usepackage{pifont}
\usepackage{natbib}                                                                                                                                  
\usepackage{microtype}
\usepackage{hyperref}
\usepackage{url}
\usepackage{booktabs}
\usepackage{graphicx}
\usepackage{xspace}
\usepackage{subcaption}
\usepackage{amsmath}
\usepackage{wrapfig}
\usepackage{longtable}
\usepackage{booktabs}
\usepackage{enumitem}
\usepackage{multirow}

\newcommand{\promptplaceholder}[1]{\texttt{\{#1\}}}
\usepackage[export]{adjustbox}

\newcommand{\ours}{\textsc{MARS}\xspace}

\newcommand{\secref}[1]{\S\ref{#1}}
\newcommand{\appref}[1]{Appendix~\ref{#1}}

\title{
Agentic Recommender System with Hierarchical Belief-State Memory
}

\affiliation{Meta Recommendation Systems (MRS)}

\author[*]{Xiang Shen}
\author[*]{Yuhang Zhou}
\author[*]{Yifan Wu}
\author{Zhuokai Zhao}
\author{Siyu Lin}
\author{Lei Huang}
\author{Qianqian Zhong}
\author{Lizhu Zhang}
\author{Benyu Zhang}
\author{Xiangjun Fan}
\author{Hong Yan}

\contribution[*]{Equal contribution}
\correspondence{\email{xiangshen@meta.com}, \email{maxfan@meta.com}}                                               

\abstract{Memory-augmented LLM agents have advanced personalized
recommendation, yet existing approaches universally adopt flat
memory representations that conflate ephemeral signals with stable
preferences, and none provides a complete lifecycle governing how
memory should evolve. We propose \ours (Memory-Augmented Agentic
Recommender System), a framework that treats recommendation as a
partially observable problem and maintains a structured belief state
that progressively abstracts noisy behavioral observations into a
compact estimate of user preferences. \ours organizes this belief
state into three tiers: event memory buffers raw signals, preference
memory maintains fine-grained mutable chunks with explicit strength
and evidence tracking, and profile memory distills all preferences
into a coherent natural language narrative. A complete lifecycle of
six operations---extraction, reinforcement, weakening,
consolidation, forgetting, and resynthesis---is adaptively scheduled
by an LLM-based planner rather than fixed-interval heuristics.
Experiments on four InstructRec benchmark domains show that \ours
achieves state-of-the-art performance with average improvements of
26.4\% in HR@1 and 10.3\% in NDCG@10 over the strongest baselines
with further gains from agentic scheduling in evolving settings.
}

\date{\today}

\begin{document}

\maketitle

\section{Introduction}
\label{sec:intro}

Recommender systems are fundamental to online content discovery,
powering personalization across social media, e-commerce, and
entertainment platforms at billion-user scale. The field has
progressed from collaborative filtering~\citep{rendle2009bpr} and
graph-based methods~\citep{he2020lightgcn} to deep learning ranking
models~\citep{naumov2019dlrm}, and more recently to self-attentive
sequential architectures~\citep{kang2018sasrec, zhai2024hstu} that
capture temporal dynamics of user behavior. While these systems
achieve strong offline metrics, they share two fundamental
limitations. First, users are passively involved, able to influence
recommendations only through implicit signals or blunt controls
rather than using natural language to directly express intent and
steer the system~\citep{carroll2025ctrlrec, zhang2023instructrec}.
Second, embedding-based models act as narrow experts that capture
statistical co-occurrences but cannot reason about \emph{why} an
item matches a user's taste~\citep{zhao2024toolrec,
peng2025survey}.

The emergence of large language models (LLMs) has opened a new
paradigm that addresses these limitations by enabling systems to
reason about user preferences and item semantics in natural
language. While LLMs can rank items
zero-shot~\citep{hou2024llmrank, lyu2024llmrec}, without persistent
memory they cannot accumulate understanding across interactions,
motivating a growing body of memory-augmented agentic recommender
systems~\citep{peng2025survey}. These systems range from static
methods that construct or rebuild user profiles from interaction
histories~\citep{xu2025iagent, shi2025personax}, to dynamic
approaches that evolve preferences incrementally through reflection
and collaborative signal
propagation~\citep{zhang2024agentcf, liao2026steam,
chen2026memrec, nguyen2026amem4rec, li2026recnet}. Despite this
progress, all existing approaches share critical limitations in how
memory is structured and managed over its lifetime.

These limitations manifest in two fundamental gaps. First, all
existing systems adopt flat memory representations that fail to
effectively perform state abstraction over the recommendation
environment. Specifically, they conflate three crucial abstraction
levels: raw, high-dimensional observations (the \emph{momentary}
level---what the user is doing now), disentangled belief variables
(the \emph{dispositional} level---what they consistently prefer,
and with what confidence), and the holistic state representation
(the \emph{identity} level---who they are as a consumer). By
collapsing these distinct tiers, a flat representation loses either
the granularity for surgical credit assignment, the recency for
detecting distribution shifts, or the semantic coherence required
for accurate ranking. Second, no existing system provides a formal
state transition mechanism---a complete memory
lifecycle---governing when and how the belief state should evolve,
leaving no principled method for extracting, consolidating, or
decaying preferences as new evidence accumulates.

To address these gaps, we propose \textbf{\ours} (Memory-Augmented
Agentic Recommender System), an agentic recommendation framework
that formulates memory management through the lens of a Partially
Observable Markov Decision
Process~\citep{kaelbling1998planning, lu2016pomdprec}. To resolve
the partial observability of underlying user intents, \ours
implements a hierarchical state abstraction architecture, assigning
each level of abstraction its own dedicated tier: \emph{event
memory} buffers raw interaction signals (the observation space),
\emph{preference memory} maintains fine-grained, independently
mutable preference chunks with explicit strength and evidence
tracking (the disentangled belief variables), and \emph{profile
memory} distills these discrete chunks into a coherent natural
language narrative (the holistic state representation). Each tier
specializes for a distinct role: events capture raw transition
dynamics, preferences enable surgical credit assignment where new
evidence updates only the relevant belief without disturbing
unrelated variables in the state space, and the profile provides
the compact, Markovian context that the LLM-based ranker consumes
most effectively.

In addition, \ours introduces the first complete memory lifecycle
for agentic recommendation, governing how information transitions
across these tiers through a principled set of operations:
extraction of preferences from raw signals, reinforcement and
weakening as new evidence accumulates, consolidation of
near-duplicate beliefs, evidence-driven forgetting, and profile
resynthesis. Unlike prior systems that either update memory after
every interaction or rebuild it at fixed intervals, an LLM planner
adaptively schedules these operations based on accumulated signal
volume, making context-aware decisions about when memory
maintenance is warranted and thereby balancing memory freshness
against computational cost.

\begin{figure}[t]
\centering
\includegraphics[width=\textwidth]{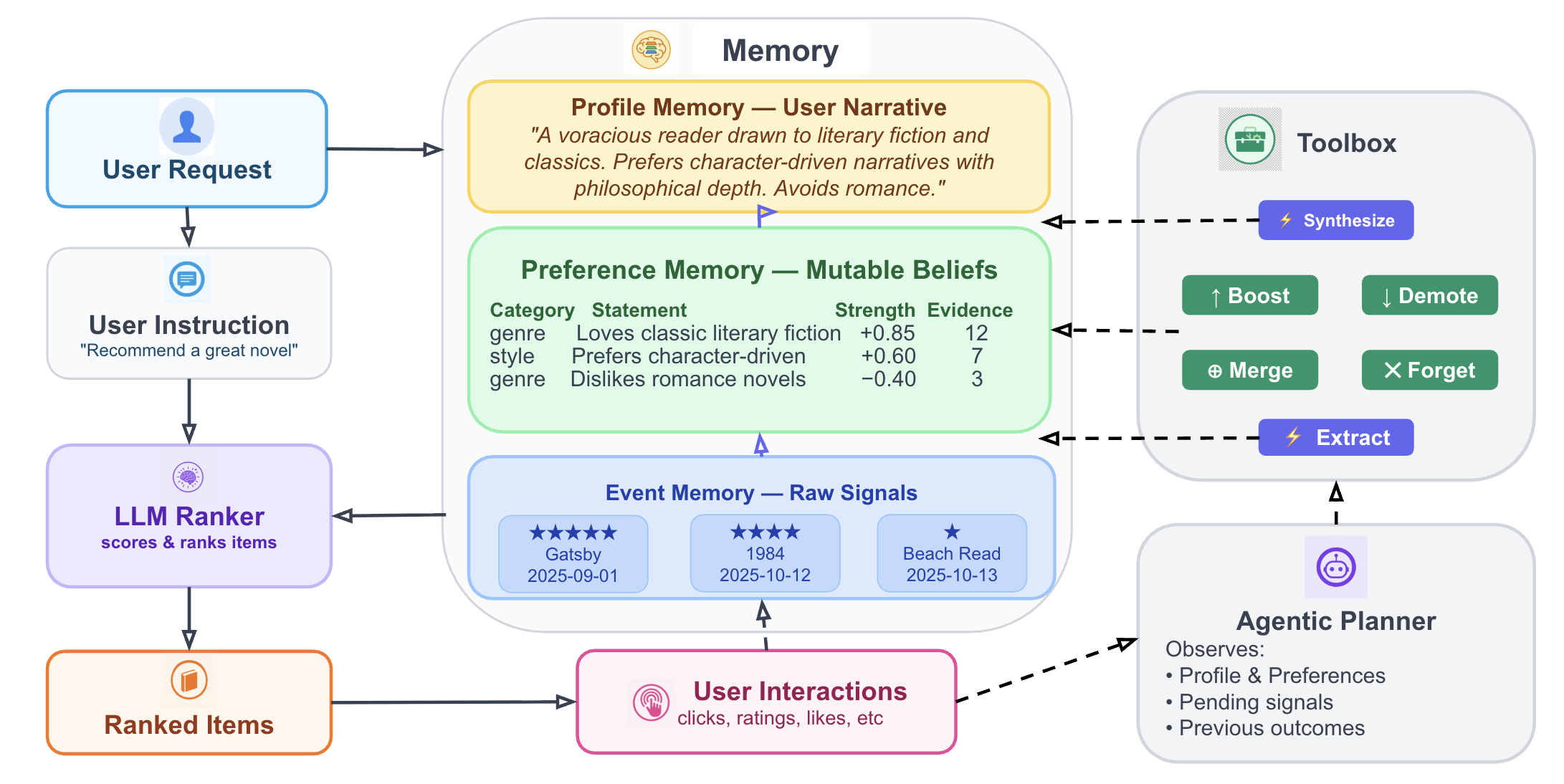}
\caption{Overview of \ours. Three-tier memory (center) stores raw
signals (event), mutable preference chunks with strength and
evidence (preference), and a synthesized user narrative (profile).
Solid arrows: online ranking path where profile and event memory
feed the LLM ranker. Dashed arrows: offline lifecycle path where
an agentic planner schedules six memory operations to keep
preferences current. Preference memory serves as an intermediate
tier that informs profile synthesis but is not directly consumed
by the ranker. The two paths are decoupled.}
\label{fig:architecture}
\end{figure}

Our primary contributions can be summarized as follows:
\begin{itemize}[leftmargin=*]
    \item We propose \textbf{\ours}, an agentic recommender system
    built around a \textbf{three-tier memory architecture} (event,
    preference, and profile memory) that resolves the ambiguity
    between short-term behavioral signals and long-term user traits
    by progressively abstracting raw events into mutable preference
    chunks and then into coherent natural language profiles.

    \item We design a \textbf{complete memory lifecycle}
    encompassing extraction, reinforcement, weakening,
    consolidation, forgetting, and resynthesis, governed by an
    LLM-based planner that adaptively schedules operations based on
    accumulated context, replacing rigid heuristics and reducing
    computational cost.

    \item We conduct \textbf{extensive experiments} on the
    InstructRec benchmark~\citep{xu2025iagent} across four domains
    in both retrospective and evolving evaluation settings,
    demonstrating state-of-the-art performance with average
    improvements of 26.4\% in HR@1 and 10.3\% in NDCG@10 over the
    strongest baseline per domain.
\end{itemize}

\section{Related Work}
\label{sec:related}

\subsection{LLM Agents in Recommendation Systems}
The application of LLM-powered agents to recommendation systems has
attracted considerable attention~\citep{peng2025survey}. One line of
work employs LLM agents to simulate user behavior for evaluation and
analysis~\citep{zhang2024agent4rec, wang2023recagent,
zhong2025ggbond, shi2025personax, bougie2025simuser,
chen2025recusersim}, demonstrating that LLM agents can accurately
reproduce user behavior but contributing primarily to evaluation
methodology rather than recommendation quality. A separate line of work
leverages LLMs as autonomous agents that orchestrate conventional
recommender models as tools~\citep{wang2024recmind,
huang2023interecagent, lei2024macrec, zhao2024toolrec, deng2025recbot,
ou2026deepresearch}. While these systems exhibit strong reasoning and
planning capabilities, they operate without persistent user memory,
treating each session independently.

The most closely related work equips LLM agents with persistent memory
to model evolving user preferences.
iAgent~\citep{xu2025iagent} introduced a user-agent-platform paradigm
that mitigates platform-side biases on behalf of the user, with i2Agent
extending it through iterative profile generation via feedback loops.
AgentCF~\citep{zhang2024agentcf} modeled both users and items as
autonomous agents that collaboratively update their memories through
forward interaction and backward reflection.
STEAM~\citep{liao2026steam} introduced atomic memory units that form
communities and prototypes, incorporating memory formation and
consolidation but lacking a forgetting mechanism. Since these per-user
approaches cannot leverage collaborative signals, more recent work
propagates preference updates across users through global memory
pools~\citep{nguyen2026amem4rec}, router-mediated
networks~\citep{li2026recnet}, or architecturally decoupled memory
managers with collaborative graph
propagation~\citep{chen2026memrec}. Despite this progress, all existing
approaches adopt flat memory representations and lack a complete
lifecycle governing extraction, consolidation, and forgetting. \ours
addresses these gaps through a three-tier architecture governed by a
full lifecycle with adaptive scheduling.

\subsection{LLM Agent Memory}
Memory management has emerged as a central challenge in LLM agent
systems. A dominant design paradigm draws inspiration from operating
systems: MemGPT~\citep{packer2023memgpt} pioneered virtual context
management with self-directed paging between main context and external
storage, and subsequent work extended this with universal memory
encapsulation and cross-type migration~\citep{li2025memos} or
heat-based tier promotion~\citep{kang2025memoryos}. Rather than
OS-level abstractions, an alternative line of work represents memory as
structured knowledge through incremental CRUD
stores~\citep{chhikara2025mem0} or temporal knowledge
graphs~\citep{rasmussen2025zep}.

Alongside architectural advances, memory evolution mechanisms have
become an active area of research. CoALA~\citep{sumers2023cognitive}
mapped cognitive science memory types onto language agents and
explicitly identified memory modification and deletion as
``understudied,'' while
MemoryBank~\citep{zhong2024memorybank} and H-Mem~\citep{ye2026h}
addressed forgetting through Ebbinghaus decay curves and feedback-driven
weight adjustment, respectively. Despite these advances, no existing
system combines extraction, consolidation, and forgetting into a unified
lifecycle; most handle forgetting through simple heuristics such as FIFO
eviction or time-based decay. Moreover, these systems are designed for
factual recall and conversational coherence rather than recommendation,
where memory must additionally capture preference intensity, evidence
strength, and multi-timescale dynamics. \ours bridges this gap by
adapting
hierarchical memory principles to preference modeling with
domain-specific lifecycle management.

\section{Methodology}\label{sec:method}

\subsection{Problem Formulation}
We consider the sequential recommendation problem with natural language
instructions. Let $\mathcal{U}$ denote the set of users and
$\mathcal{I}$ denote the set of items. Given a user $u \in \mathcal{U}$
with historical interactions
$\mathcal{H}_u = \{(i_1, t_1), (i_2, t_2), \dots, (i_T, t_T)\}$
ordered by timestamp where each $i_k \in \mathcal{I}$, an optional
natural language instruction $q_u$ expressing the user's current intent,
and a candidate set $\mathcal{C} \subseteq \mathcal{I}$ of items to
rank, the goal is to produce a ranking of $\mathcal{C}$ that maximizes
relevance to the user's preferences and intent.

To model the uncertainty of user intent, we draw on the POMDP
framework~\citep{kaelbling1998planning, lu2016pomdprec}. The
user's true preferences constitute a hidden state
$s \in \mathcal{S}_{\text{hidden}}$ governed by unknown transition
dynamics $\mathcal{T}(s' \mid s, a)$; the system cannot observe
$s$ directly but receives partial, noisy observations
$o \in \Omega$ through behavioral signals (clicks, purchases,
ratings). In a standard POMDP, an agent maintains a probabilistic
belief state updated via Bayesian filtering, but this is
intractable when preferences live in open-ended natural language.
\ours instead maintains a structured, symbolic belief state
$M_u = (\mathcal{E}_u, \mathcal{P}_u, \mathcal{S}_u)$ that
progressively abstracts raw observations ($\mathcal{E}_u$) into
discrete, disentangled belief variables ($\mathcal{P}_u$) and a
holistic semantic representation ($\mathcal{S}_u$). An LLM-based
planner $\pi_{\text{plan}}(a_t \mid M_u)$ selects discrete
memory-updating actions $a_t \in \mathcal{A}$ (e.g., extract,
boost, demote, merge, forget, synthesize), balancing memory freshness against
computational cost, while a ranking policy $\pi_{\text{rank}}$
generates the item list from $\mathcal{C}$ conditioned on $M_u$
and the user instruction $q_u$.

\subsection{Three-Tier Memory Architecture}
Following our POMDP formulation, \ours constructs its symbolic
belief state $M_u$ through three tiers of increasing abstraction,
as illustrated in Figure~\ref{fig:architecture}.

\paragraph{\textbf{Event Memory.}}
The first tier
$\mathcal{E}_u = \{e_1, e_2, \dots, e_L\}$ captures the momentary
level, storing raw behavioral signals as structured records. Each
signal $e = (u, i, a, m_i, t)$ captures the user $u$, item $i$,
action type $a$, item metadata $m_i$, and timestamp $t$. In our
framework, this tier serves as the raw observation space $\Omega$,
faithfully recording behavioral data ($o \in \Omega$) without
interpretation. Event memory is maintained as a bounded FIFO queue
with capacity $L_{\max}$. We denote by
$\mathcal{E}_u^{\text{pending}} \subseteq \mathcal{E}_u$ the
subset of signals that have not yet been processed by preference
extraction; once processed, observations remain in
$\mathcal{E}_u$ as ranking context but are no longer pending.

\paragraph{\textbf{Preference Memory.}}
The second tier
$\mathcal{P}_u = \{p_1, p_2, \dots, p_K\}$ captures the
dispositional level, representing the user's preferences as
fine-grained, independently mutable natural language statements
acting as disentangled belief variables. Each preference chunk $p$
is a tuple of four fields: a domain-specific $\text{category}$
label (e.g., genre, writing style), a natural language
$\text{statement}$ describing the preference, a
$\text{strength} \in [-1, 1]$ score indicating intensity and
polarity, and an $\text{evidence} \in \mathbb{N}$ count tracking
how many interactions have reinforced or contradicted this belief
(see Figure~\ref{fig:architecture} for an example). These strength
and evidence fields act as explicit confidence intervals over the
belief state. Preference chunks are constructed by an LLM-based
extractor $\pi_{\text{ext}}$ that acts as an
observation-to-belief mapping, analyzing pending event signals
against existing preferences to produce structured updates:
$$\Delta\mathcal{P} =
  \pi_{\text{ext}}(\mathcal{E}_u^{\text{pending}}, \mathcal{P}_u)$$
Because each chunk is independently mutable, new evidence updates
only the relevant belief dimensions rather than globally rewriting
the user representation, enabling precise credit assignment.
Preference memory feeds directly into profile memory, where the
full set of discrete variables is consolidated into a cohesive
narrative.

\paragraph{\textbf{Profile Memory.}}
The third tier $\mathcal{S}_u$ captures the identity level,
distilling all preference chunks into a coherent natural language
narrative of the user's overall taste profile. The profile is
generated by an LLM-based synthesizer $\pi_{\text{syn}}$:
$$\mathcal{S}_u =
  \pi_{\text{syn}}(\mathcal{P}_u, \mathcal{S}_u^{\text{prev}})$$
where $\mathcal{S}_u^{\text{prev}}$ is the previous profile
version, included to maintain narrative continuity across state
updates. Beyond aggregating individual preference chunks, the
synthesized profile captures relationships between preferences and
resolves potential tensions, producing the most compressed,
Markovian state representation that the ranker consumes alongside
recent behavioral signals.

This three-tier design offers key advantages over flat
representations. Profile resynthesis is amortized over multiple
mutations rather than triggered after every interaction, yielding
computational savings. Both preference and profile tiers are
expressed in natural language, making the belief state fully
interpretable and providing a semantically rich input for LLM-based
ranking. However, without proactive pruning and consolidation, the
preference tier will inevitably accumulate redundant or
contradicted beliefs, introducing high-dimensional noise. This
motivates a formal transition mechanism to govern state
maintenance.

\subsection{Memory Lifecycle Management}
The memory lifecycle governs the transition dynamics
$\mathcal{T}(s' \mid s, a)$ of the belief state $M_u$ as new
observations arrive. It defines six primitive operations that
constitute the discrete action space $\mathcal{A}$ exposed to an
LLM-based planner $\pi_{\text{plan}}$, which acts as a transition
policy selecting actions based on the current belief state.

\paragraph{\textbf{Memory Operations.}}
Five operations act on preference memory $\mathcal{P}_u$ and one
on profile memory $\mathcal{S}_u$. The preference operations serve
as deterministic, heuristic alternatives to exact Bayesian belief
updates:
\begin{itemize}[leftmargin=*, nosep]
    \item \textbf{Extract}: invokes $\pi_{\text{ext}}$ to create
    new preference chunks from pending signals. In evolving mode,
    this is the only tool requiring an LLM call; the remaining
    operations are deterministic state updates.
    \item \textbf{Boost}: reinforces a chunk by incrementing its
    strength and evidence count, increasing state confidence.
    \item \textbf{Demote}: weakens a chunk by decrementing
    strength. The demote step $\delta_-$ is larger than the boost
    step $\delta_+$, implementing a pessimistic update rule where
    contradictory evidence erodes beliefs faster than confirmatory
    evidence reinforces them.
    \item \textbf{Merge}: consolidates near-duplicate chunks
    identified by the planner, directly reducing state collinearity
    and redundancy.
    \item \textbf{Forget}: prunes chunks whose strength has decayed
    below a threshold after sufficient evidence, acting as an
    explicit decay function that removes stale or contradicted
    state variables.
    \item \textbf{Synthesize}: regenerates $\mathcal{S}_u$ from
    $\mathcal{P}_u$ via $\pi_{\text{syn}}$. Also triggered
    automatically every $\gamma$ mutations as a safeguard.
\end{itemize}
After any mutation sequence, a per-category capacity constraint
retains only the top $K_{\text{cat}}$ chunks ranked by
$\text{evidence}_p \cdot |\text{strength}_p|$, mathematically
bounding the dimensionality of the belief state.

\paragraph{\textbf{Agentic Scheduling.}}
Rather than triggering operations on a fixed schedule, the planner
$\pi_{\text{plan}}$ acts as a frozen heuristic policy over the
belief state, selecting which actions $a_t \in \mathcal{A}$ to
apply:
$$a_t = \pi_{\text{plan}}(\mathcal{S}_u, \mathcal{P}_u,
  \mathcal{E}_u^{\text{pending}}, H_{\text{prev}})$$
where $H_{\text{prev}}$ records outcomes of the previous round.
The planner receives the current belief state---profile, preference
chunks with confidence metadata, pending observations, and a
memory health summary---and outputs either an empty action list
(skip, the default when new observations are strictly consistent
with existing beliefs) or an ordered sequence of transitions
$a_t = [(t_1, \text{params}_1), \dots,
(t_m, \text{params}_m)]$ over the plannable operations. The skip
action reflects an implicit reward optimization: when new signals
carry no marginal information, the expected advantage of a state
transition is negative, meaning the computational cost outweighs
the utility. In practice, the policy invokes extraction when
observations suggest novel latent intents, schedules merge when
state redundancy is detected, and triggers forgetting when belief
strength decays below optimal thresholds. The full procedure is
detailed in Algorithm~\ref{alg:lifecycle}
(\appref{app:algorithm}).

\subsection{LLM-based Ranking}
Given the abstracted belief state $M_u$ and candidate set
$\mathcal{C}$, \ours produces a ranked list through pointwise
scoring. The ranking prompt incorporates two forms of user context:
the holistic state representation $\mathcal{S}_u$ and recent
behavioral observations from $\mathcal{E}_u$. Preference memory
$\mathcal{P}_u$ is not directly included in the ranking prompt;
instead, its contribution is mediated through profile synthesis,
which distills the discrete variables into the coherent narrative
consumed by the ranker. This design avoids redundancy between raw
preference chunks and the profile that already summarizes them, as
validated by the ablation study in \secref{sec:ablation}. When a
natural language instruction $q_u$ is provided, it receives
explicit priority over historical beliefs. All candidates are
presented in a single prompt, and the ranking policy assigns each
an independent relevance score along with a brief rationale:
$$\{r_1, \dots, r_N\} =
  \pi_{\text{rank}}(\mathcal{C}, \mathcal{S}_u,
  \mathcal{E}_u, q_u)$$
When $|\mathcal{C}|$ exceeds the batch size, candidates are
partitioned into batches scored independently. The final ranking is
obtained by sorting candidates in descending order of $r_j$.

\section{Experiments}\label{sec:experiments}

\subsection{Experiment Setup}

\paragraph{\textbf{Datasets.}}
We evaluate on four domains from the InstructRec
benchmark~\citep{xu2025iagent}, which augments Amazon product
reviews~\citep{ni2019justifying} (\textbf{Books}, \textbf{MovieTV}),
Goodreads~\citep{wan2019spoiler} (\textbf{Goodreads}), and
\textbf{Yelp}~\citep{yelp2024dataset} with natural language
instructions. The four domains span a range of sparsity levels and
interaction patterns, as summarized in Table~\ref{tab:datasets}.
Retrospective experiments are conducted on all four domains with
full user sets, while evolving experiments use two Books subsamples
(100 users each) due to the high per-signal LLM computational
cost:
\emph{Active User} and \emph{Light User}, allowing us to study how
evolving memory updates perform across users with rich versus limited
history.

\begin{table}[h]
\centering
\caption{Dataset statistics. Density = Interactions /
(Users $\times$ Items).}
\label{tab:datasets}
\begin{tabular}{lccccc}
\toprule
\textbf{Dataset} & \textbf{Users} & \textbf{Items} & \textbf{Interactions} & \textbf{Density} & \textbf{Avg/User} \\ \midrule
Books            & 7,377          & 120,925        & 207,759               & 0.023\%          & 28.2              \\
Goodreads        & 11,734         & 57,364         & 618,330               & 0.092\%          & 52.7              \\
MovieTV          & 5,649          & 28,987         & 79,737                & 0.049\%          & 14.1              \\
Yelp             & 2,950          & 31,636         & 63,142                & 0.068\%          & 21.4              \\ \midrule
Books (Active User)   & 100            & 8,334          & 8,935                 & 1.072\%          & 89.3              \\
Books (Light User)   & 100            & 1,505          & 1,534                 & 1.019\%          & 15.3              \\ \bottomrule
\end{tabular}
\vspace{-0.15in}
\end{table}

\paragraph{\textbf{Evaluation Modes.}}
To comprehensively validate the proposed approach, we evaluate
under two complementary settings. \emph{Retrospective} mode follows
the standard protocol used by prior
work~\citep{chen2026memrec, xu2025iagent}: the full interaction
history $\mathcal{H}_u$ is processed at a single point in time to
build the memory state $M_u$. Concretely, the extraction step
ingests all interactions to construct preference memory
$\mathcal{P}_u$ and synthesize profile memory $\mathcal{S}_u$,
while event memory $\mathcal{E}_u$ retains only the most recent
$L_{\max}$ signals as ranking context. This mode enables direct
comparison with baselines under identical conditions.
\emph{Evolving} mode further validates the memory lifecycle by
simulating online operation, where memory evolves continuously as
behavioral signals arrive sequentially. We evaluate two scheduling
strategies: \emph{fixed}, which triggers memory operations every
$B$ signals, and \emph{agentic}, where $\pi_{\text{plan}}$
adaptively decides when updates are warranted. Unless otherwise
stated, we report results in retrospective mode.

\paragraph{\textbf{Evaluation Metrics.}}
Following prior work~\citep{chen2026memrec, xu2025iagent}, we use a
leave-one-out evaluation protocol where the most recent interaction per
user is held out for testing. Each test instance consists of 1
ground-truth item and 9 randomly sampled negative items ($N=10$
candidates). We report HR@1, HR@5, NDCG@5, and NDCG@10.

\paragraph{\textbf{Baselines.}}
We compare against both traditional recommendation models and
LLM-based recommendation agents:
\begin{itemize}
    \item \textbf{BPR}~\citep{rendle2009bpr}: Bayesian personalized
    ranking with matrix factorization, optimizing pairwise item
    preferences.
    \item \textbf{LightGCN}~\citep{he2020lightgcn}: Graph collaborative
    filtering with linear embedding propagation over the user-item
    interaction graph.
    \item \textbf{SASRec}~\citep{kang2018sasrec}: Self-attentive
    sequential model that captures long-term dependencies in user
    interaction sequences.
    \item \textbf{LLMRank}~\citep{hou2024llmrank}: Zero-shot LLM
    ranking without persistent user memory.
    \item \textbf{AgentCF}~\citep{zhang2024agentcf}: User and item
    agents with collaborative memory updated through pairwise
    interaction and reflection.
    \item \textbf{iAgent / i2Agent}~\citep{xu2025iagent}: User-agent
    paradigm with static (iAgent) and iterative (i2Agent) profile
    generation via feedback loops.
    \item \textbf{MemRec}~\citep{chen2026memrec}: Decoupled memory
    manager and reasoning LLM with collaborative memory graph
    propagation.
\end{itemize}

\paragraph{\textbf{Implementation Details.}}
Traditional baselines are trained using
RecBole~\citep{zhao2021recbole}. All LLM-based methods use
Llama-4-Maverick-17B-128E-Instruct~\citep{meta2025llama4} as the
default backbone LLM unless otherwise specified (see LLM Backbone
ablation in \secref{sec:ablation}). Following
i2Agent~\citep{xu2025iagent}, all memory-based approaches use a
history window of $L_{\max} = 15$ recent interactions (full
hyperparameters in Appendix~\ref{app:implementation}).

\subsection{Retrospective Mode Results}
Table~\ref{tab:retro} presents the retrospective results across all
four InstructRec domains.

\begin{table}[t]
\centering
\footnotesize
\setlength{\tabcolsep}{2pt}
\caption{Retrospective mode results across four domains. Best in
\textbf{bold}, second best \underline{underlined}. H, N denote HR,
NDCG. Improv.\ shows \% gain of \ours over second best.}
\label{tab:retro}
\begin{tabular}{l cccc cccc cccc cccc}
\toprule
& \multicolumn{4}{c}{\textbf{Books}} & \multicolumn{4}{c}{\textbf{Goodreads}} & \multicolumn{4}{c}{\textbf{MovieTV}} & \multicolumn{4}{c}{\textbf{Yelp}} \\
\cmidrule(lr){2-5} \cmidrule(lr){6-9} \cmidrule(lr){10-13} \cmidrule(lr){14-17}
\textbf{Model} & H@1 & H@5 & N@5 & N@10 & H@1 & H@5 & N@5 & N@10 & H@1 & H@5 & N@5 & N@10 & H@1 & H@5 & N@5 & N@10 \\
\midrule
BPR      & .140 & .563 & .340 & .483 & \underline{.341} & .722 & \underline{.539} & \underline{.627} & .263 & .622 & .441 & .562 & .231 & .666 & .444 & .553 \\
LightGCN & .152 & .574 & .343 & .486 & .285 & .640 & .466 & .580 & .330 & .671 & .499 & .607 & .313 & \underline{.817} & .562 & .624 \\
SASRec   & .198 & .707 & .464 & .559 & .234 & .685 & .462 & .563 & .188 & .483 & .337 & .500 & .198 & .454 & .332 & .497 \\
\midrule
LLMRank  & .215 & .602 & .416 & .539 & .218 & .603 & .417 & .540 & .294 & .716 & .513 & .601 & .265 & .520 & .398 & .546 \\
AgentCF  & .263 & .620 & .445 & .562 & .124 & .380 & .251 & .438 & .171 & .365 & .271 & .458 & .150 & .479 & .315 & .474 \\
iAgent   & .335 & .664 & .499 & .605 & .162 & .593 & .377 & .505 & .380 & .749 & .572 & .650 & .289 & .681 & .481 & .582 \\
i2Agent  & .325 & .716 & .526 & .613 & .164 & .662 & .417 & .522 & .349 & .768 & .569 & .640 & .259 & .673 & .468 & .569 \\
MemRec   & \underline{.396} & \underline{.770} & \underline{.591} & \underline{.662} & .257 & \underline{.741} & .500 & .584 & \underline{.471} & \underline{.850} & \underline{.668} & \underline{.715} & \underline{.402} & .757 & \underline{.590} & \underline{.665} \\
\midrule
\textbf{\ours} & \textbf{.499} & \textbf{.846} & \textbf{.682} & \textbf{.731} & \textbf{.369} & \textbf{.770} & \textbf{.568} & \textbf{.643} & \textbf{.591} & \textbf{.896} & \textbf{.752} & \textbf{.785} & \textbf{.587} & \textbf{.887} & \textbf{.752} & \textbf{.787} \\
\rowcolor{gray!10}
\textit{Improv.} & \textit{+26.0} & \textit{+9.9} & \textit{+15.4} & \textit{+10.4} & \textit{+8.2} & \textit{+3.9} & \textit{+5.4} & \textit{+2.6} & \textit{+25.5} & \textit{+5.4} & \textit{+12.6} & \textit{+9.8} & \textit{+46.0} & \textit{+8.6} & \textit{+27.5} & \textit{+18.3} \\
\bottomrule
\end{tabular}
\vspace{-0.15in}
\end{table}

\ours achieves the best performance on all four metrics across all
four domains, as shown in the Improv.\ rows. HR@1 gains range from
8.2\% (Goodreads) to 46.0\% (Yelp), with improvements consistently
larger on HR@1 than on deeper-rank metrics, indicating that the
three-tier memory architecture helps the ranker precisely identify
the single most relevant item by providing fine-grained preference
resolution.

Among LLM-based methods, MemRec is the strongest competitor,
ranking second on Books and MovieTV across most metrics. Notably,
traditional CF baselines remain competitive in denser domains: BPR
is second-best on three of four metrics on Goodreads, the densest
dataset (0.092\% density), and LightGCN achieves second-best HR@5
on Yelp, suggesting that dense collaborative signals partially
compensate for the absence of semantic reasoning. Nevertheless,
\ours consistently outperforms both categories by combining
structured memory with language-based ranking.

\subsection{Evolving Mode Results}
To validate the memory lifecycle under online operation, we
evaluate on the Books Active User and Light User subsamples in
evolving mode. We compare against AgentCF, MemRec, and i2Agent,
the approaches that support or can be adapted to evolving memory
maintenance. Table~\ref{tab:evolving} presents results for these
baselines and two \ours scheduling strategies: \emph{fixed}, which
triggers memory operations every $B$ signals, and \emph{agentic},
where $\pi_{\text{plan}}$ adaptively decides when updates are
warranted. Both \ours variants incorporate user instructions at
ranking time when available.

\begin{table}[t]
\centering
\footnotesize
\setlength{\tabcolsep}{3.5pt}
\caption{Evolving mode results on Books. Best in \textbf{bold}, second
best \underline{underlined}.}
\label{tab:evolving}
\begin{tabular}{l cccc cccc}
\toprule
& \multicolumn{4}{c}{\textbf{Active User}} & \multicolumn{4}{c}{\textbf{Light User}} \\
\cmidrule(lr){2-5} \cmidrule(lr){6-9}
\textbf{Model} & H@1 & H@5 & N@5 & N@10 & H@1 & H@5 & N@5 & N@10 \\
\midrule
AgentCF          & .230 & .650 & .443 & .554 & .340 & .640 & .488 & .604 \\
i2Agent          & .490 & .750 & \underline{.635} & \underline{.717} & \underline{.600} & .850 & \underline{.729} & \underline{.776} \\
MemRec           & .360 & .730 & .554 & .638 & .450 & .800 & .636 & .697 \\
\ours (fixed)    & \underline{.500} & \underline{.760} & .632 & .710 & .510 & \underline{.860} & .692 & .736 \\
\textbf{\ours (agentic)} & \textbf{.510} & \textbf{.830} & \textbf{.678} & \textbf{.733} & \textbf{.620} & \textbf{.930} & \textbf{.784} & \textbf{.807} \\
\rowcolor{gray!10}
\textit{Improv.} & \textit{+2.0} & \textit{+9.2} & \textit{+6.8} & \textit{+2.2} & \textit{+3.3} & \textit{+8.1} & \textit{+7.5} & \textit{+4.0} \\
\bottomrule
\end{tabular}
\end{table}

\ours (agentic) achieves the best results across all metrics on both
user groups, while \ours (fixed) remains competitive, surpassing all
baselines on Active Users and matching i2Agent on Light Users. The
agentic advantage over fixed scheduling is substantially larger on
Light Users (+21.6\% HR@1) than Active Users (+2.0\%), because limited
history makes each scheduling decision more impactful.
A detailed case study tracing how the planner uses demote and forget to handle a preference shift for a single user is provided in \appref{app:case_study}.

\begin{figure}[t]
\centering
\includegraphics[width=0.55\linewidth]{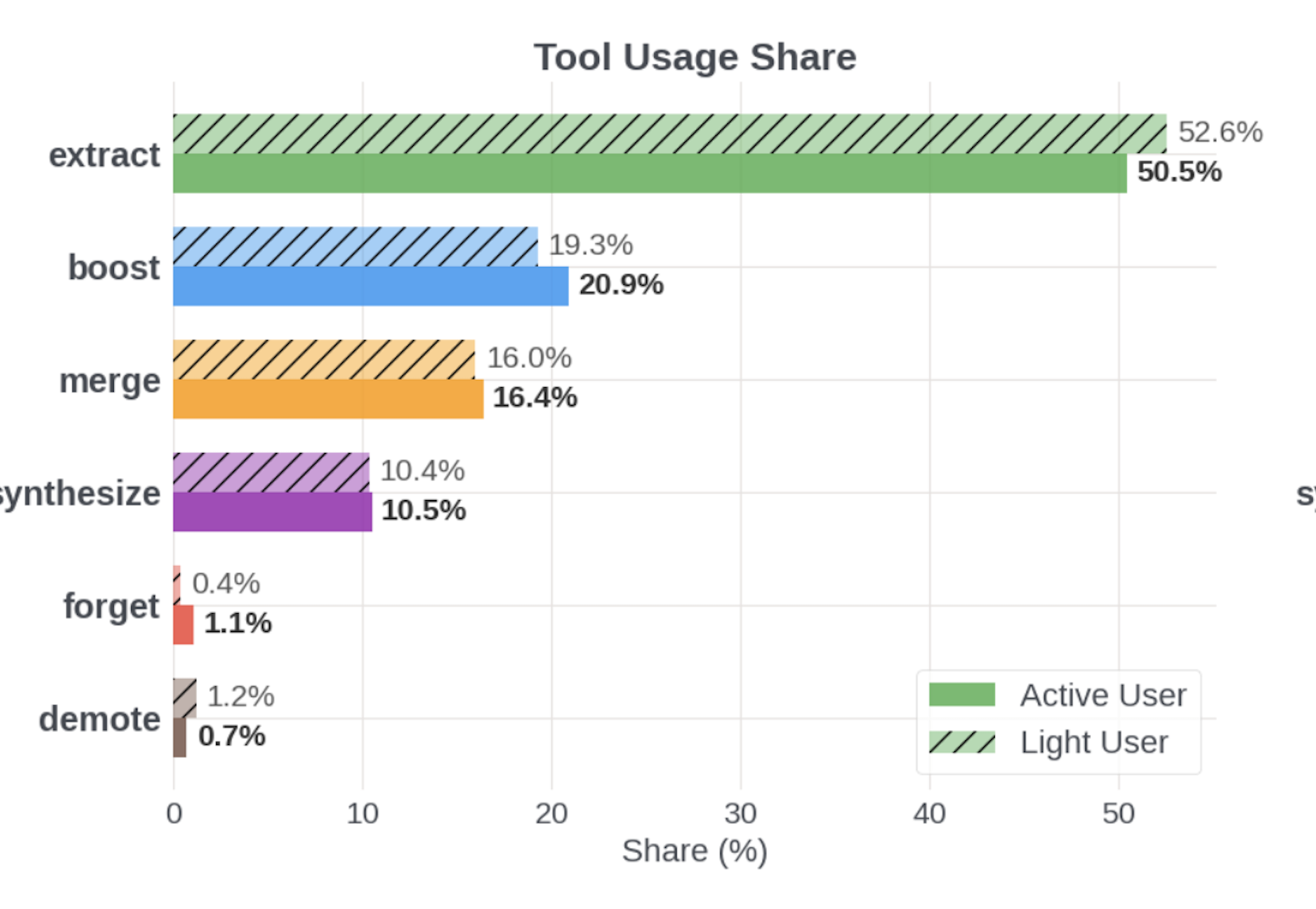}
\caption{Planner tool usage distribution on Books: extract and boost
dominate ($\sim$70\%), while forget and demote account for $<$2\%.}
\label{fig:tool_usage}
\vspace{-0.15in}
\end{figure}

\subsection{Ablation Studies}\label{sec:ablation}

\paragraph{\textbf{Memory Architecture Tiers.}}

\begin{figure}[t]
\centering
\includegraphics[width=0.8\linewidth]{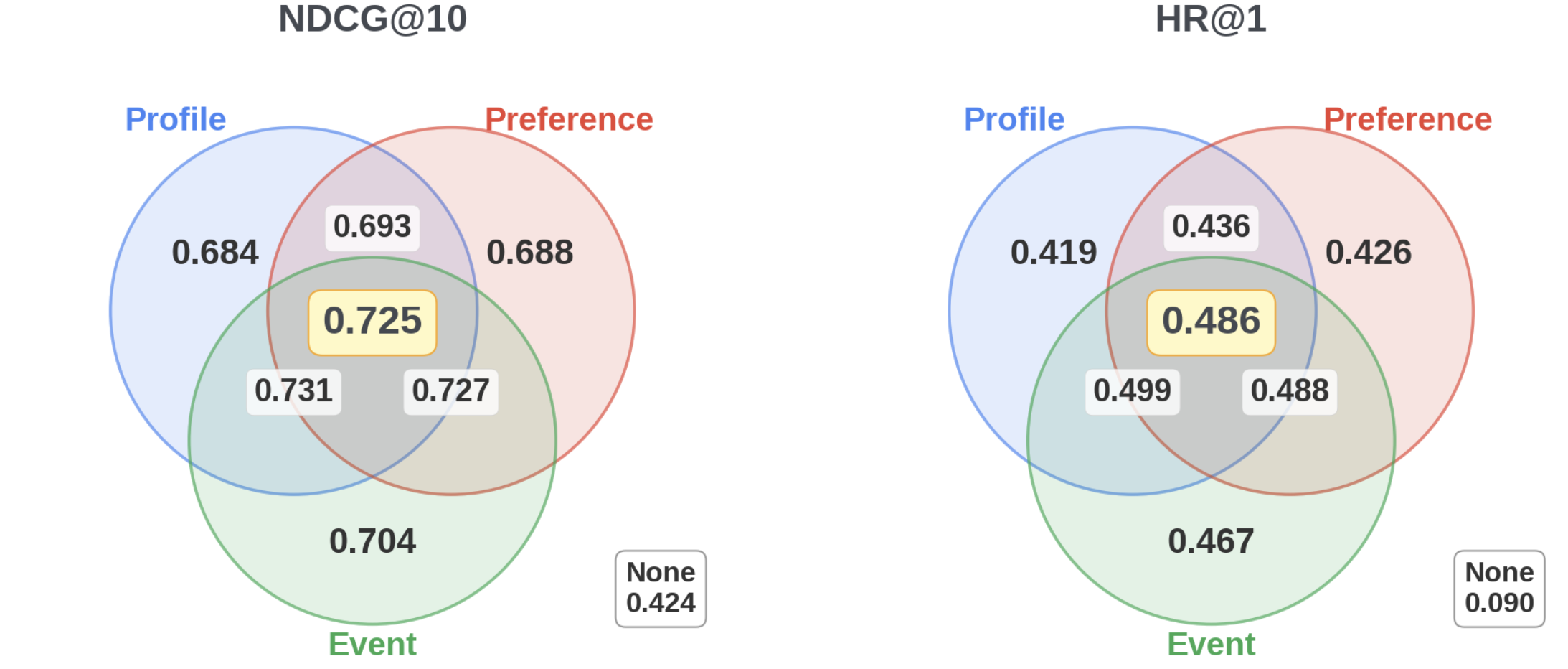}
\caption{Memory tier ablation on Books (7,377 users). Left: NDCG@10;
right: HR@1. Profile$+$event (0.731) achieves the best NDCG@10.}
\label{fig:memory_venn}
\vspace{-0.15in}
\end{figure}

We evaluate the contribution of each memory tier through a full
factorial study ($2^3 = 8$ combinations of $\pm$event,
$\pm$preference, $\pm$profile memory) on the Books dataset with all
7,377 users (Figure~\ref{fig:memory_venn}).
Two observations emerge.
\textbf{(1) Any memory dramatically outperforms none, and
profile$+$event achieves the best results.}
The no-memory baseline (NDCG@10 = 0.424) underperforms every
single-tier configuration by $>$60\% relative, confirming that
persistent memory is essential. The profile$+$event combination
achieves the highest NDCG@10 (0.731), validating our design choice
of feeding only profile and event memory to the ranker while using
preference memory as an intermediate lifecycle tier.
\textbf{(2) Preference memory contributes through profile synthesis,
not direct ranking input.}
Adding preference chunks directly to the ranking prompt alongside
profile and event memory (all three tiers: 0.725) slightly
underperforms profile$+$event alone (0.731). This confirms that
preference memory's value lies in structuring the lifecycle --- enabling
fine-grained extraction, boosting, and forgetting that produce a
higher-quality profile --- rather than as a direct signal for
ranking, where it introduces redundancy with the profile that
already summarizes it.

\paragraph{\textbf{LLM Backbone.}}
We study the effect of varying the planner LLM in evolving
agentic mode while fixing ranking and memory to Maverick-17B
(Table~\ref{tab:ablation_planner}). We compare three models
spanning different scales:
Llama-4-Maverick-17B-128E-Instruct~\citep{meta2025llama4} (the
default), Llama-3.1-8B-Instruct (a smaller, faster alternative),
and Gemini-3-Flash~\citep{google2026gemini3}.

\begin{table}[t]
\centering
\footnotesize
\setlength{\tabcolsep}{3pt}
\caption{Planner LLM ablation in evolving agentic mode on Books.
Ranking and memory fixed to Maverick-17B; only the planner LLM varies.
Best in \textbf{bold} within each sample group.}
\label{tab:ablation_planner}
\begin{tabular}{l cccc cccc}
\toprule
& \multicolumn{4}{c}{\textbf{Active User}} & \multicolumn{4}{c}{\textbf{Light User}} \\
\cmidrule(lr){2-5} \cmidrule(lr){6-9}
\textbf{Planner} & H@1 & H@5 & N@5 & N@10 & H@1 & H@5 & N@5 & N@10 \\
\midrule
Llama-4-17B & .510 & .830 & .678 & .733 & .620 & .930 & .784 & .807 \\
Llama-3-8B        & .490 & .790 & .651 & .719 & .650 & .930 & .793 & .815 \\
Gemini-3-Flash    & \textbf{.540} & \textbf{.840} & \textbf{.698} & \textbf{.750} & \textbf{.680} & \textbf{.950} & \textbf{.814} & \textbf{.830} \\
\bottomrule
\end{tabular}
\end{table}

Gemini-3-Flash achieves the best results as a planner on both
Active Users (HR@1=0.540, +5.9\% over Maverick) and Light Users
(HR@1=0.680, +9.7\%), despite being a smaller model. This
indicates that planning and ranking require different LLM
capabilities: planning benefits from fast, decisive
instruction-following, while ranking requires deeper semantic
reasoning over item descriptions and user preferences.
Llama-3.1-8B as planner also surpasses Maverick on Light Users
(0.650 vs.\ 0.620), suggesting that a smaller, faster planner can
be equally or more effective, with practical computational cost
implications for deployment.

\begin{figure}[t]
\centering
\includegraphics[width=0.95\linewidth]{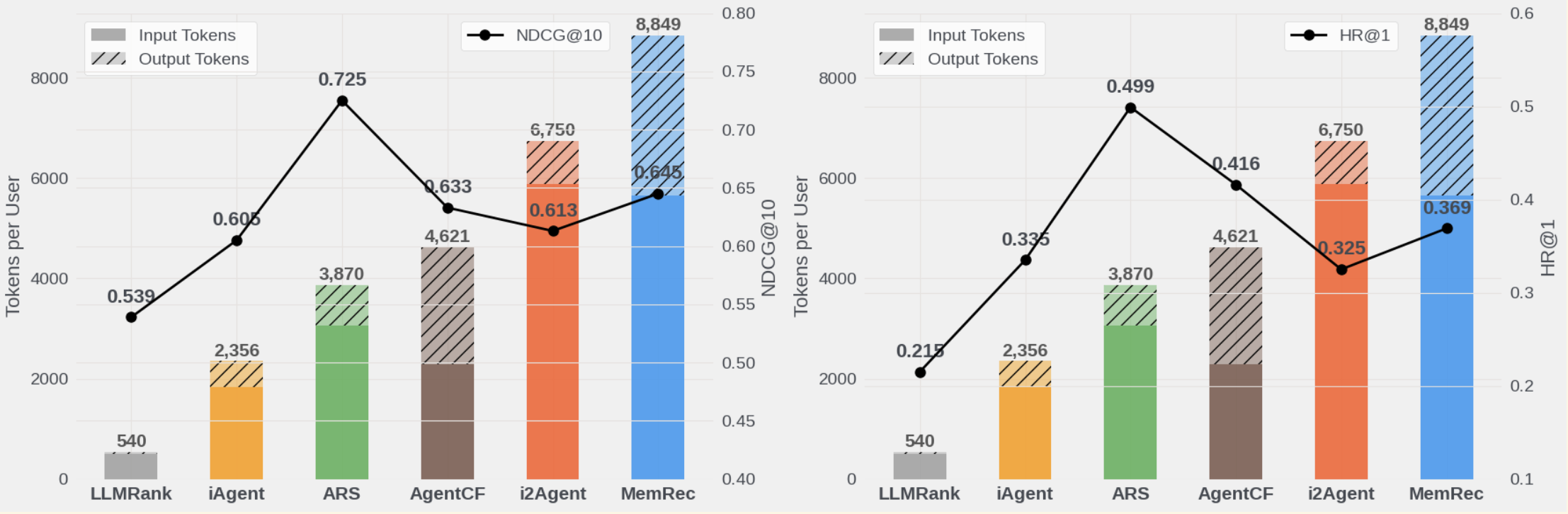}
\caption{Computational cost--quality tradeoff (Books,
Maverick-17B). Bars: per-user tokens (input + output). Line:
recommendation quality.}
\label{fig:cost}
\vspace{-0.15in}
\end{figure}

\subsection{Computational Cost Analysis}
We analyze computational efficiency by measuring per-user LLM token
consumption across all approaches on Books using Llama-4-Maverick-17B.
Figure~\ref{fig:cost} shows the computational cost--quality
tradeoff: stacked bars decompose total tokens into input (prompt)
and output (completion), while the overlaid line tracks
recommendation quality.

\ours uses only 3 LLM calls per user totaling 3,870 tokens,
compared to MemRec's 6 calls and 8,849 tokens---a 2.3$\times$
reduction achieved by compressing interaction history into compact
preference chunks and replacing multi-stage pipelines with a
single ranking call.

\section{Conclusion}
\label{sec:conclusion}

We presented \ours, an agentic recommendation framework that maintains
a three-tier belief state (event, preference, profile) through an
LLM-governed lifecycle of six operations. Experiments across four
domains show 26.4\% average HR@1 and 10.3\% NDCG@10 gains over the
strongest baselines at 2.3$\times$ fewer tokens, with agentic
scheduling yielding up to 21.6\% further gains in evolving settings.
The ablation reveals that preference memory's value lies not in
directly informing the ranker, but in structuring the lifecycle
operations that produce a higher-quality profile --- acting as the
intermediate state representation through which noisy observations
are refined into a compact belief state. This suggests that in
partially observable preference modeling, the quality of the state
abstraction pipeline matters more than the richness of the state
fed to the decision-maker.

\clearpage
\bibliographystyle{plainnat}
\bibliography{ars}

\clearpage

\appendix

\section{Agentic Memory Update Algorithm}\label{app:algorithm}

\begin{algorithm}[h]
\caption{Agentic Memory Update}\label{alg:lifecycle}
\begin{algorithmic}[1]
\REQUIRE User $u$, interaction stream $\mathcal{H}_u$, memory
state $M_u = (\mathcal{E}_u, \mathcal{P}_u, \mathcal{S}_u)$
\REQUIRE Planner check interval $B$, synthesis threshold $\gamma$
\STATE $\text{mutations} \gets 0$
\FOR{each new behavioral signal $e$}
    \STATE $\mathcal{E}_u \gets \mathcal{E}_u \cup \{e\}$
    \hfill $\triangleright$ Append to event memory
    \IF{$|\mathcal{E}_u^{\text{pending}}| \geq B$}
        \STATE $a_t \gets \pi_{\text{plan}}(\mathcal{S}_u,
        \mathcal{P}_u, \mathcal{E}_u^{\text{pending}},
        H_{\text{prev}})$ \hfill $\triangleright$ Schedule ops
        \FOR{each $(t_k, \text{params}_k) \in a_t$}
            \STATE Execute operation $t_k$ on $M_u$
            \STATE $\text{mutations} \gets \text{mutations} + 1$
        \ENDFOR
        \STATE Apply per-category capacity constraint to
        $\mathcal{P}_u$
        \IF{$\text{mutations} \geq \gamma$}
            \STATE $\mathcal{S}_u \gets
            \pi_{\text{syn}}(\mathcal{P}_u, \mathcal{S}_u)$
            \hfill $\triangleright$ Auto-synthesize
            \STATE $\text{mutations} \gets 0$
        \ENDIF
    \ENDIF
\ENDFOR
\end{algorithmic}
\end{algorithm}

\section{Supplemental Implementation Details}\label{app:implementation}

\paragraph{\textbf{\ours Hyperparameters.}}
Table~\ref{tab:hyperparams} summarizes the full set of hyperparameters
used for \ours.

\begin{table}[h]
\centering
\caption{\ours hyperparameter settings.}
\label{tab:hyperparams}
\begin{tabular}{lll}
\toprule
\textbf{Parameter} & \textbf{Symbol} & \textbf{Value} \\ \midrule
Planner check interval & $B$ & 3 \\
Max preferences per category & $K_{\text{cat}}$ & 8 \\
Event memory window & $L_{\max}$ & 15 \\
Boost step & $\delta_+$ & 0.1 \\
Demote step & $\delta_-$ & 0.2 \\
Forgetting strength threshold & $\epsilon$ & 0.1 \\
Forgetting evidence threshold & $n_{\min}$ & 5 \\
Synthesis trigger & $\gamma$ & 5 mutations \\
Candidate slate size & $N$ & 10 \\
Concurrent evaluation batch & -- & 32 users \\
\bottomrule
\end{tabular}
\end{table}

\paragraph{\textbf{Decoding Configuration.}}
All LLM calls across all methods use temperature 0 (greedy
decoding) to ensure deterministic outputs and minimize
stochasticity across runs. No sampling, top-$k$, or nucleus
filtering is applied.

\paragraph{\textbf{User Subsample Selection.}}
Evolving mode experiments use two 100-user subsamples from the Books
domain, selected by interaction count and a fixed random seed for
reproducibility. \emph{Active Users}: users with 50--200 interactions,
sampled with seed 42. \emph{Light Users}: users with 10--30
interactions, sampled with seed 64. The same user sets are used
consistently across all evolving experiments and ablations to ensure
comparability.

\paragraph{\textbf{Preference Categories.}}
Each domain uses six preference categories that structure the
preference memory tier. Categories are generated once per domain by
prompting the LLM with 10 sample item descriptions and asking it
to identify the most discriminative axes of preference.
Table~\ref{tab:categories} lists the resulting categories.

\begin{table}[h]
\centering
\footnotesize
\caption{Preference categories per domain, generated by a one-time LLM
call on 10 sample item descriptions.}
\label{tab:categories}
\begin{tabular}{ll}
\toprule
\textbf{Domain} & \textbf{Categories} \\
\midrule
Books & genre, writing style, theme, setting, author type, mood \\
Goodreads & genre, writing style, theme, series pref., mood, character type \\
MovieTV & genre, mood, era, theme, quality, pacing \\
Yelp & cuisine, price range, ambiance, occasion, location, service \\
\bottomrule
\end{tabular}
\end{table}

\section{Additional Ablation Studies}\label{app:ablation}

\paragraph{\textbf{Memory Capacity.}}
Table~\ref{tab:capacity} shows the effect of varying the maximum
preferences per category $K_{\text{cat}}$ on Books (all users).
Performance is stable across
$K_{\text{cat}} \in \{2, 4, 8, 16\}$, with HR@1 ranging from
0.486 to 0.499 and NDCG@10 from 0.725 to 0.731. This robustness
suggests that the consolidation and forgetting mechanisms
effectively manage preference quality regardless of capacity, and
that a small number of high-confidence preferences per category
suffices for ranking. In summary, preference capacity is not a
sensitive hyperparameter for \ours, and practitioners can set
$K_{\text{cat}}$ without careful tuning.

\begin{table}[h]
\centering
\footnotesize
\caption{Effect of preference capacity $K_{\text{cat}}$ on Books (all
users). Performance is stable across all settings.}
\label{tab:capacity}
\begin{tabular}{l cccc}
\toprule
$K_{\text{cat}}$ & H@1 & H@5 & N@5 & N@10 \\
\midrule
2  & .490 & .844 & .679 & .728 \\
4  & .486 & .845 & .676 & .725 \\
8 (default) & .499 & .846 & .682 & .731 \\
16 & .487 & .845 & .677 & .726 \\
\bottomrule
\end{tabular}
\end{table}

\section{Computational Cost Estimation}\label{app:cost}

Table~\ref{tab:cost_estimate} estimates the total LLM token consumption
and USD computational cost for \ours across all evaluation
settings, using Gemini 2.5 Flash pricing (\$0.30/M input,
\$2.50/M output) as a reference.

\begin{table}[h]
\centering
\footnotesize
\caption{Estimated \ours token consumption and USD computational
cost. USD computed using Gemini 2.5 Flash pricing (\$0.30/M input,
\$2.50/M output); costs scale linearly with model pricing tier.}
\label{tab:cost_estimate}
\begin{tabular}{llrrrr}
\toprule
\textbf{Mode} & \textbf{Domain} & \textbf{Users} & \textbf{Tokens/User} & \textbf{Total (M)} & \textbf{Est.\ USD} \\
\midrule
Retrospective & Books     & 7,377  & 3,870  & 28.5 & \$21.75 \\
Retrospective & Goodreads & 11,734 & 3,870  & 45.4 & \$34.74 \\
Retrospective & MovieTV   & 5,649  & 3,870  & 21.9 & \$16.69 \\
Retrospective & Yelp      & 2,950  & 3,870  & 11.4 & \$8.70 \\
\midrule
Evolving & Active User & 100 & $\sim$68,500 & 6.8 & \$5.12 \\
Evolving & Light User  & 100 & $\sim$12,800 & 1.3 & \$1.05 \\
\midrule
\multicolumn{2}{l}{\textbf{Total}} & 27,910 & -- & 115.3 & \textbf{\$88.05} \\
\bottomrule
\end{tabular}
\end{table}

At Gemini 2.5 Flash pricing (\$0.30/M input tokens, \$2.50/M
output tokens), the total computational cost for all \ours
experiments is approximately \$88. Traditional baselines (BPR,
LightGCN, SASRec) are trained with RecBole on a single GPU with
negligible computational cost relative to LLM inference.

\section{Limitations}\label{app:limitations}

\paragraph{\textbf{Scope of Evaluation.}}
This work focuses on LLM-native recommendation with natural language
instructions available, a setting where structured memory and
language-based reasoning provide the greatest advantage. In this
setting, \ours consistently outperforms both traditional collaborative
filtering models and LLM-based baselines. However, on dense domains
where collaborative signals are strong (e.g., Goodreads), traditional
models such as BPR remain competitive, suggesting that the current
framework is most beneficial when semantic reasoning over instructions
and item descriptions plays a central role.

\paragraph{\textbf{Evolving Evaluation Scale and Variance.}}
Evolving mode experiments use 100-user subsamples due to the high
per-signal LLM computational cost of sequential memory updates.
This scale is consistent with prior work in the same
setting~\citep{zhang2024agentcf, nguyen2026amem4rec}, which
evaluate on comparable or smaller user populations. We do not
report error bars; however, all LLM calls use temperature 0
(greedy decoding) to minimize stochasticity. Residual variance
arises only from non-deterministic infrastructure factors such as
GPU kernel scheduling, floating-point accumulation order, and
batched request routing, which are negligible in practice. The
memory capacity ablation (Table~\ref{tab:capacity}) corroborates
this: performance varies by at most 0.013 in HR@1 and 0.006 in
NDCG@10 across four hyperparameter settings, indicating high
stability.

\paragraph{\textbf{Absence of Collaborative Signals.}}
\ours operates on a per-user basis and does not propagate preference
updates across users. While collaborative signals have been shown to
benefit recommendation in prior
work~\citep{chen2026memrec, nguyen2026amem4rec, li2026recnet}, the
three-tier architecture is designed to be modular: collaborative
memory components (e.g., shared preference graphs, cross-user
memory propagation) can be integrated as additional tools available
to the planner without modifying the core lifecycle. Exploring this
integration is an important direction for future work.

\section{Case Study: Memory Lifecycle in Action}\label{app:case_study}

We illustrate the memory lifecycle on a representative user from
the Books evolving evaluation (User~1773, 65 training
interactions, HR@1\,=\,1.0). This user is a theology reader whose
interests briefly shift toward historical biographies before
returning to their core domain. The agentic planner uses five of
six lifecycle operations---extract, boost, demote, forget, and
synthesize---to track this preference evolution.

\paragraph{\textbf{Phase 1: Discovery (interactions 1--30).}}
The planner extracts core preferences: topics such as
``Christianity'' (0.8), ``spiritual formation'' (0.7),
``reformed theology'' (0.8), and authors including
``Dietrich Bonhoeffer'' (0.9) and ``Michael Horton'' (0.8).
By step 15, the user has 11 preference chunks across two
categories.

\paragraph{\textbf{Phase 2: Preference Shift (interactions 30--40).}}
The user reads books by Churchill and Lincoln, which the planner
extracts as new author preferences (strength 0.7 each). However,
subsequent interactions return to theology. The planner recognizes
these as tangential and issues two rounds of \textbf{demote}
operations:
\begin{itemize}[nosep,leftmargin=*]
\item Round 1: demote ``Churchill'' ($0.7 \to 0.5$) and
``Lincoln'' ($0.7 \to 0.5$), while simultaneously extracting
``postmodern christianity'' and ``Henri Nouwen''.
\item Round 2: demote ``Churchill'' ($0.5 \to 0.3$) and
``Lincoln'' ($0.5 \to 0.3$), while extracting ``immigration''
and ``Douglas Murray''.
\end{itemize}
The preference count dips from 15 to 14 at this stage, reflecting
the planner's active curation.

\paragraph{\textbf{Phase 3: Pruning and Reinforcement
(interactions 40--50).}}
The demoted preferences are now weak enough to trigger
\textbf{forget}: the planner removes ``Churchill'' and ``Lincoln''
entirely, while simultaneously \textbf{boosting} ``reformed
theology'' ($0.8 \to 0.9 \to 1.0$) and ``John Calvin''
($0.8 \to 0.9 \to 1.0$). Profile \textbf{resynthesis} fires
automatically after every 5 mutations, producing updated
narratives that reflect the refined preference state.

\paragraph{\textbf{Phase 4: Continued Exploration
(interactions 50--65).}}
The planner continues extracting niche theology topics
(``covenant theology'', ``lord's supper'', ``beatitudes'',
``Heidelberg Catechism'') and periodically boosting
``reformed theology''. The preference count stabilizes at 16.

\paragraph{\textbf{Result.}}
The final memory state contains 16 preferences across 2
categories (topic, author), with 5 profile resyntheses performed
over the session. The ranker, consuming the synthesized profile
and recent event signals, ranks the ground-truth test item first
(HR@1\,=\,1.0). The lifecycle tool distribution for this user
(extract 52\%, boost 23\%, synthesize 11\%, demote 9\%,
forget 5\%) illustrates all three phases of memory
management: discovery, belief revision, and consolidation.

\section{Prompt Templates}\label{app:prompts}

This section presents the core LLM prompt templates used in \ours.
Variable placeholders are shown in \promptplaceholder{braces}.

\begin{tcolorbox}[colback=gray!3, colframe=gray!50,
  title={\textbf{Prompt 1: Preference Extraction}
  ($\pi_{\text{ext}}$)}, fonttitle=\small,
  fontupper=\scriptsize, left=4pt, right=4pt, top=2pt,
  bottom=2pt]
\textit{System:} You are analyzing user engagement to understand
their preferences.\\[4pt]
Given these recent engagements with
\promptplaceholder{item\_noun}:\\
\promptplaceholder{engagements\_table}\\[2pt]
Current known preferences:\\
\promptplaceholder{existing\_preferences}\\[4pt]
Extract or update preference statements. For each preference:\\
1. \textbf{category}: \promptplaceholder{category\_list}\\
2. \textbf{text}: Natural language statement (be specific, not
generic)\\
3. \textbf{strength}: $-$1.0 (strong dislike) to 1.0 (strong
like)\\
4. \textbf{action}: ``create'' $|$ ``strengthen'' $|$
``weaken''\\[4pt]
Rules:\\
-- Be specific: ``Enjoys cerebral sci-fi with philosophical
themes'' $>$ ``Likes sci-fi''\\
-- Look for patterns: 3 similar engagements $\to$ preference,
1 $\to$ too early\\
-- Maximum 5 preferences per response\\[2pt]
Output as JSON array only, no other text.
\end{tcolorbox}

\begin{tcolorbox}[colback=gray!3, colframe=gray!50,
  title={\textbf{Prompt 2: Profile Synthesis}
  ($\pi_{\text{syn}}$)}, fonttitle=\small,
  fontupper=\scriptsize, left=4pt, right=4pt, top=2pt,
  bottom=2pt]
\textit{System:} You are a user preference analyst. Write concise,
coherent profiles.\\[4pt]
Group these preferences into a coherent 150--300 word profile
paragraph that captures this user's taste. Write in third person
(``This user\ldots'').\\[4pt]
Preference statements:\\
\promptplaceholder{chunks\_text}\\[4pt]
Previous profile (maintain continuity):\\
\promptplaceholder{previous\_profile}\\[4pt]
Write a single coherent paragraph (150--300 words). Do NOT use
bullet points. Output ONLY the profile paragraph.
\end{tcolorbox}

\begin{tcolorbox}[colback=gray!3, colframe=gray!50,
  title={\textbf{Prompt 3: Agentic Scheduler}
  ($\pi_{\text{plan}}$)}, fonttitle=\small,
  fontupper=\scriptsize, left=4pt, right=4pt, top=2pt,
  bottom=2pt]
\textit{System:} You are a memory manager for a recommendation
system. Analyze preferences against new items. Output ONLY valid
JSON.\\[4pt]
\textbf{User Profile:} \promptplaceholder{full\_profile}\\
\textbf{Current Preferences}
(\promptplaceholder{pref\_count} chunks):
\promptplaceholder{preference\_list}\\
\textbf{New Items} (\promptplaceholder{pending\_count} pending):
\promptplaceholder{items\_with\_descriptions}\\
\textbf{Memory Health:} \promptplaceholder{health\_section}\\[4pt]
\textbf{Available Actions:}\\
1. \texttt{extract} -- Create new preferences\\
2. \texttt{merge} -- Merge similar preferences\\
3. \texttt{boost} -- Reinforce a preference\\
4. \texttt{demote} -- Weaken a preference\\
5. \texttt{forget} -- Delete a preference\\[4pt]
\textbf{Guidelines:}\\
-- \textbf{Skip is the default.} If new items match existing
preferences, output \texttt{\{``actions'': []\}}\\
-- Only extract when items reveal genuinely NEW interests\\
-- Be selective: at most 3 actions per response\\[2pt]
Output: \texttt{\{``actions'': [\ldots]\}} or
\texttt{\{``actions'': []\}}
\end{tcolorbox}

\begin{tcolorbox}[colback=gray!3, colframe=gray!50,
  title={\textbf{Prompt 4: LLM Ranking}
  ($\pi_{\text{rank}}$)}, fonttitle=\small,
  fontupper=\scriptsize, left=4pt, right=4pt, top=2pt,
  bottom=2pt]
You are a recommendation engine. Score how well each item matches
this user's preferences.\\[4pt]
\promptplaceholder{user\_profile\_section}\\
\promptplaceholder{session\_memory\_section}\\
\promptplaceholder{instruction\_section}\\[4pt]
\textbf{Items to Score:}\\
\promptplaceholder{items\_table}\\[4pt]
\textbf{Instructions:} Score each item on a 0--10 scale:\\
-- 8--10: \textsc{strong} match -- high confidence user will enjoy\\
-- 4--7: \textsc{maybe} -- moderate match\\
-- 0--3: \textsc{weak} -- poor match\\[4pt]
Use the full 0--10 range. Spread scores apart. Best match should
score 9+, worst should score 2 or below.\\[4pt]
\textbf{Output Format} (one line per item, no extra text):\\
\texttt{ITEM\_ID | SCORE | TIER | brief\_reason}
\end{tcolorbox}

\end{document}